\documentclass{article} 
\usepackage{iclr16_multi,times}
\usepackage{hyperref}
\usepackage{url}

\title{Multi-task Sequence to Sequence Learning}
\author{Minh-Thang Luong\thanks{Minh-Thang Luong is also a student at Stanford
University.}, Quoc V. Le, Ilya Sutskever, Oriol Vinyals, Lukasz Kaiser \\
Google Brain\\
\texttt{lmthang@stanford.edu,\{qvl,ilyasu,vinyals,lukaszkaiser\}@google.com} \\
}

\iclrfinalcopy 

\usepackage{caption} 
\usepackage{subcaption} 
\usepackage{multirow} 
\usepackage{amsmath} 
\usepackage{amssymb} 
\usepackage{graphicx}
\usepackage{color} 
\usepackage{pstricks} 
\usepackage{arydshln} 

\newcommand{\imgExt}{eps}

\newcommand{\hide}[1]{}

\newcommand{\tgt}[1]{y_{#1}} 
\newcommand{\src}[1]{x_{#1}} 

\newcommand{\MB}[1]{\mbox{\boldmath{$#1$}}} 
\newcommand{\open}[1]{\left(#1\right)} 
\newcommand{\eq}[1]{Eq.~(\ref{#1})}

\newcommand{\ssl}{\emph{seq2seq}} 
\newcommand{\otm}{one-to-many}
\newcommand{\mto}{many-to-one}
\newcommand{\mtm}{many-to-many}

\begin{document}

\maketitle

\begin{abstract}
  Sequence to sequence learning has recently emerged as a new
  paradigm in supervised learning.
  To date, most of its applications focused on only one task and not much work
  explored this framework for multiple tasks.  This paper
  examines three multi-task learning (MTL) settings for sequence to sequence
  models:
  (a) the {\it one-to-many} setting -- where the encoder is shared
  between several tasks such as machine translation and
  syntactic parsing, (b) the {\it many-to-one} setting -- useful when only the
  decoder can be shared, as in the case of 
  translation and image caption generation, and (c) the {\it
    many-to-many} setting -- where multiple encoders and decoders are
  shared, which is the case with unsupervised objectives
  and translation.  Our results show that training on a small amount of parsing and
  image caption data can improve the translation quality between English and
  German by up to $1.5$ BLEU
  points over strong single-task baselines on the WMT benchmarks. Furthermore,
  we have established a new {\it
  state-of-the-art} result in constituent parsing with 93.0 F$_1$. Lastly, we reveal interesting properties of the two unsupervised learning
  objectives, autoencoder and skip-thought, in the MTL context: autoencoder helps less in terms of
  perplexities but more on BLEU scores compared to skip-thought.
\end{abstract}

\section{Introduction}
\label{sec:intro}
Multi-task learning (MTL) is an important machine learning paradigm that
aims at improving the generalization performance of a task using other related
tasks. 
Such framework has been widely studied by
\citet{thrun96,caruana97,evgeniou04,ando05,argyriou07,kumar12}, among many
others. In the context of deep neural networks, MTL has
been applied successfully to various problems ranging from language
\citep{liu15}, to vision
\citep{donahue14},
and speech \citep{heigold13,huang2013cross}.

Recently, sequence to sequence (\ssl{}) learning, proposed by
\citet{kal13}, \citet{sutskever14}, and \citet{cho14}, emerges as an effective paradigm for dealing with
variable-length inputs and outputs. \ssl{} learning, at its core, uses
recurrent neural networks to map variable-length input sequences to
variable-length output sequences.  While relatively new, the \ssl{}
approach has achieved state-of-the-art results in not only its original
application -- machine translation --
\citep{luong15,jean15,luong15attn,jean15wmt,luong15iwslt}, but also image caption generation \citep{vinyals15caption},
and constituency parsing \citep{vinyals15grammar}. 

Despite the popularity of multi-task learning and sequence to sequence
learning, there has been little work in combining MTL with \ssl{}
learning. To the best of our knowledge, there is only one recent
publication by \citet{dong15} which applies a \ssl{} models for machine
translation, where the goal is to translate from one language to
multiple languages.
In this work, we propose three MTL
approaches that complement one another: (a) the {\it \otm} approach -- for
tasks that can have an encoder in common, such as translation and parsing; this 
applies to the multi-target translation setting in \citep{dong15} as well, (b)
the {\it \mto} approach -- useful for multi-source
translation or tasks in which only the decoder can be easily shared,
such as translation and image captioning, and lastly, (c) the {\it \mtm} approach -- which share
multiple encoders and decoders through which we study the effect of unsupervised
learning in translation.
We show
that syntactic parsing and image caption generation improves the
translation quality between English and German by up to +$1.5$ BLEU points over
strong single-task baselines on the WMT benchmarks. 
Furthermore, we have established a new {\it state-of-the-art} result in
constituent parsing with 93.0 F$_1$.
We also explore two unsupervised learning
objectives, sequence autoencoders \citep{dai15} and skip-thought vectors
\citep{kiros15skip}, and reveal their interesting properties in the MTL setting: autoencoder helps less in terms of
  perplexities but more on BLEU scores compared to skip-thought.

\section{Sequence to Sequence Learning}
\label{sec:seq2seq}
\begin{figure}
\centering
\includegraphics[width=1\textwidth, clip=true, trim= 0 0 0
0]{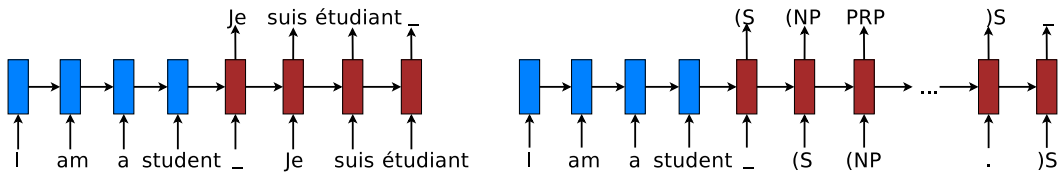}
\caption{{\bf Sequence to sequence learning examples} -- (left) machine
translation \citep{sutskever14} and ({\it right}) constituent parsing
\citep{vinyals15grammar}.}
\label{f:s2s}
\end{figure}

Sequence to sequence learning (\ssl{}) aims to directly model the conditional
probability $p(\tgt{}|\src{})$ of mapping an input sequence,
$\src{1},\ldots,\src{n}$, into an output sequence, $\tgt{1},\ldots,\tgt{m}$.
It accomplishes such goal through the {\it encoder-decoder} framework proposed
by \citet{sutskever14} and \citet{cho14}. As illustrated in Figure~\ref{f:s2s},
the {\it encoder} computes a representation $\MB{s}$
for each input sequence. Based on that input representation,
the {\it decoder} generates an output sequence, one unit at a time, and hence, decomposes the conditional probability as:
\begin{equation}
\log p(\tgt{}|\src{}) = \sum_{j=1}^m \nolimits \log
p\open{\tgt{j}|\tgt{<j},\src{},\MB{s}}
\label{e:s2s}
\end{equation}

A natural model for sequential data is the recurrent
neural network (RNN), which is used by most of the recent \ssl{} work.
These work,
however, differ in terms of: (a) {\it architecture} -- from unidirectional, to
bidirectional, and deep multi-layer RNNs; and (b) {\it RNN type} -- which are long-short term memory (LSTM)
\citep{lstm97} and the gated recurrent unit \citep{cho14}. 

Another important difference between \ssl{} work lies in what constitutes the
input representation $\MB{s}$.
The early \ssl{} work \citep{sutskever14,cho14,luong15,vinyals15caption} uses only the last encoder state
 to initialize the decoder and sets $\MB{s}\!=\![\text{ }]$ in 
\eq{e:s2s}. Recently, \citet{bog15} proposes an {\it attention mechanism}, a way
to provide \ssl{} models with a random access memory, to 
handle long input sequences.
This is accomplished by setting
$\MB{s}$ in \eq{e:s2s} to be the set of encoder hidden states already computed. On the decoder side, at each time step, the attention mechanism will
decide how much information to retrieve from that memory by learning where to
focus, i.e., computing the alignment weights for all input positions. Recent work such as \citep{xu15,jean15,luong15attn,vinyals15grammar}
has found that it is crucial to empower \ssl{} models with the attention mechanism.

\section{Multi-task Sequence-to-Sequence Learning}
\label{sec:multi}
We generalize the work of \citet{dong15} to the multi-task sequence-to-sequence
learning setting that includes the tasks of machine translation (MT),
constituency parsing, and image caption generation. Depending which tasks 
involved, we propose to categorize multi-task \ssl{} learning into three general
settings.
In addition, we will discuss the unsupervised learning tasks considered as well
as the learning process.

\subsection{One-to-Many Setting}
\label{subsec:otm}
\begin{figure}
\centering
\includegraphics[width=0.5\textwidth, clip=true, trim= 0 0 0
0]{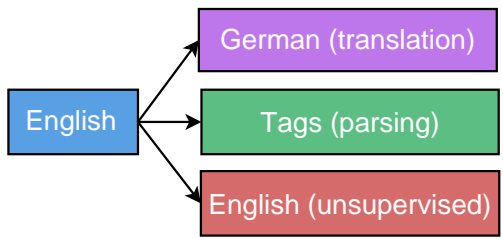}
\caption{{\bf One-to-many Setting} -- one encoder, multiple decoders. This scheme
is useful for either multi-target translation as
in \cite{dong15} or between different tasks. Here, English and
German imply sequences of words in the respective languages. The $\alpha$ values
give the proportions of parameter updates that are allocated for the different tasks.
} 
\label{f:otm}
\end{figure}

This scheme involves {\it one encoder} and {\it multiple decoders} for tasks in
which the encoder can be shared, as illustrated in
Figure~\ref{f:otm}. The input to each task is a sequence of
English words. A separate decoder is used to generate each sequence of
output units which can be either (a) a sequence of tags for
constituency parsing as used in \citep{vinyals15grammar}, (b) a
sequence of German words for machine translation \citep{luong15attn},
and (c) the same sequence of English words for autoencoders or a
related sequence of English words for the skip-thought objective
\citep{kiros15skip}.

\subsection{Many-to-One Setting}
\label{subsec:mto}
\begin{figure}[tbh]
\centering
\includegraphics[width=0.55\textwidth, clip=true, trim= 0 0 0
0]{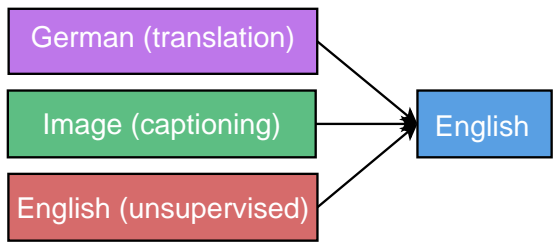}
\caption{{\bf Many-to-one setting} -- multiple encoders, one decoder. This scheme
is handy for tasks in which only the decoders can be shared.}
\label{f:mto}
\end{figure}

This scheme is the opposite of the {\it one-to-many}
setting. As illustrated in Figure~\ref{f:mto}, it consists of {\it multiple
encoders} and {\it one decoder}. This is useful for tasks in which only the
decoder can be shared, for example, when our tasks include machine translation
and image caption generation \citep{vinyals15caption}. In addition, from a machine
translation perspective, this setting can benefit from a large
amount of monolingual data on the target side, which is a standard
practice in machine translation system and has also been explored
for neural MT by \cite{gulcehre2015using}.

\subsection{Many-to-Many Setting}
\label{subsec:mtm}
\begin{figure}
\centering
\includegraphics[width=0.75\textwidth, clip=true, trim= 0 0 0
0]{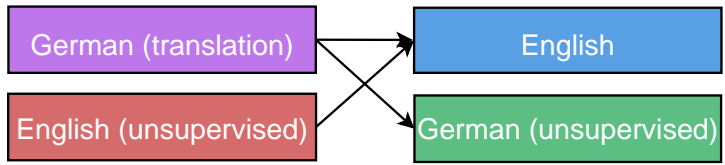}
\caption{{\bf Many-to-many setting} -- multiple encoders, multiple decoders. We
consider this scheme in a limited context of machine translation to utilize the large
monolingual corpora in both the source and the target languages. Here, we
consider a single translation task and two unsupervised autoencoder tasks.} 
\label{f:mtm}
\end{figure}

Lastly, as the name describes, this category is the most general one,
consisting of multiple encoders and multiple decoders.
We will explore this scheme in a translation setting that involves sharing multiple
encoders and multiple decoders.  In addition to the machine
translation task, we will include two unsupervised 
objectives over the source and target languages as illustrated in
Figure~\ref{f:mtm}.

\subsection{Unsupervised Learning Tasks}

Our very first unsupervised learning task involves learning {\it autoencoders} from
monolingual corpora, which has recently been applied to sequence to sequence
learning \citep{dai15}. However, in \citet{dai15}'s work, the authors
only experiment with pretraining and then finetuning, but not joint training which
can be viewed as a form of multi-task learning (MTL). As such, we are
very interested in knowing whether the same trend extends to our MTL settings.

Additionally, we investigate the use of the {\it skip-thought}
vectors \citep{kiros15skip} in the context of our MTL framework.
Skip-thought vectors are trained by training sequence to sequence
models on pairs of consecutive sentences, which makes the skip-thought
objective a natural \ssl{} learning candidate. A minor technical
difficulty with skip-thought objective is that 
the training data must consist of ordered sentences, e.g., paragraphs.  Unfortunately, in
many applications that include machine translation, we only have
sentence-level data where the sentences are unordered. To
address that, we split each sentence into two halves; we then use 
one half to predict the other half.
\subsection{Learning}
\label{subsec:learning}
\cite{dong15} adopted an {\it alternating} training approach, where they
optimize each task for a fixed number of parameter updates (or
mini-batches) before switching to the next task (which is a different
language pair). In our setting, our tasks are more diverse and contain
different amounts of training data. As a result, we allocate different
numbers of parameter updates for each task, which are expressed with
the {\it mixing} ratio values $\alpha_i$ (for each task $i$). Each
parameter update consists of training data from one task only. When
switching between tasks, we select randomly a new task $i$ with
probability $\frac{\alpha_i}{\sum_j \alpha_j}$.

Our convention is that the first task is the
{\it reference} task with $\alpha_1 = 1.0$ and the number of training
parameter updates for that task is prespecified to be $N$. A typical task $i$ will then be
trained for $\frac{\alpha_i}{\alpha_1}\cdot N$ parameter updates.
Such convention makes it easier for us to fairly compare the same reference
task in a single-task setting which has also been trained for exactly $N$
parameter updates.

When sharing an encoder or a decoder, we share both the recurrent connections
and the corresponding embeddings.


\section{Experiments}
\label{sec:exp}
We evaluate the multi-task learning setup on a wide variety of
sequence-to-sequence tasks: constituency parsing, image caption
generation, machine translation, and a number of unsupervised learning as
summarized in Table~\ref{t:tasks}.

\subsection{Data}
\label{subsec:data}
Our experiments are centered around the {\it translation} task, where we aim to determine 
whether other tasks can improve translation and vice versa. We use the WMT'15 data
\citep{bojar15} for the English$\leftrightarrows$German
translation problem. Following 
\citet{luong15attn}, we use the 50K most frequent words for each
language from the training corpus.\footnote{The corpus has already been tokenized using the default
tokenizer from Moses.  Words not in these vocabularies are represented by the token
\texttt{<unk>}.} These vocabularies are then shared with other tasks, except for
parsing in which the target ``language'' has a vocabulary of 104 tags. 
We use newstest2013 (3000 sentences) as a validation set to select our
hyperparameters, e.g., mixing coefficients. For testing, to be comparable with existing results in
\citep{luong15attn}, we use the filtered
newstest2014 (2737
sentences)\footnote{\url{http://statmt.org/wmt14/test-filtered.tgz}} for the
English$\rightarrow$German translation task and newstest2015 (2169
sentences)\footnote{\url{http://statmt.org/wmt15/test.tgz}}
for the German$\rightarrow$English task.
See the summary in Table~\ref{t:tasks}.

For the {\it unsupervised} tasks, we use the English and German monolingual corpora
from WMT'15.\footnote{The training sizes reported for
the unsupervised tasks are
only 10\% of
the original WMT'15 monolingual corpora which we randomly sample from. Such reduced sizes are
for faster training time and already about three times larger than that of the parallel
data. We consider using all the monolingual data in future work.} Since in
our experiments, unsupervised tasks are always coupled with translation tasks,
we use the same validation and test sets as the accompanied translation tasks.

For {\it constituency parsing}, we experiment with two types of corpora:
\begin{enumerate}
\item a small corpus -- the widely used
Penn Tree Bank (PTB) dataset \citep{Marcus:1993:BLA} and,
\item a large corpus -- the high-confidence (HC) parse trees 
provided by \citet{vinyals15grammar}.
\end{enumerate}
The two parsing tasks, however, are evaluated on the same validation (section
22) and test (section 23)
sets from the PTB data. Note also that the parse trees have been linearized
following \citet{vinyals15grammar}. 
Lastly, for {\it image caption generation}, we use a dataset of image and caption pairs provided by
\citet{vinyals15caption}.

\subsection{Training Details}

In all experiments, following \citet{sutskever14} and \citet{luong15}, we train deep LSTM
models as follows: (a) we use 4 LSTM layers each of which has
1000-dimensional cells and embeddings,\footnote{For image caption generation, we use 1024
dimensions, which is also the size of the image embeddings.} (b) parameters are
uniformly initialized in [-0.06, 0.06], (c) we use a mini-batch size of 128, (d)
dropout is applied with probability of 0.2 over vertical connections
\citep{pham2014dropout}, (e) we use SGD with a fixed
learning rate of 0.7, (f) input sequences are reversed, and lastly, (g) we use a simple finetuning schedule -- after $x$
epochs, we halve the learning rate every $y$ epochs. The values $x$ and $y$
are referred as {\it finetune start} and {\it finetune cycle} in
Table~\ref{t:tasks} together with the number of training epochs per task.

As described in Section~\ref{sec:multi}, for each multi-task
experiment, we need to choose one task to be the {\it reference
task} (which corresponds to $\alpha_1 = 1$). The choice of the
reference task helps specify the number of training epochs and the
finetune start/cycle values which we also when training that reference
task alone for fair comparison. To make sure our findings are
reliable, we run each experimental configuration twice and
report the average performance in the format {\it mean (stddev)}.

\begin{table}
\centering
\resizebox{14cm}{!}{
\begin{tabular}{l|c|c|c|c|c|c|c|c}
\multirow{ 2}{*}{\bf{Task}} & {\bf Train} & {\bf Valid} &{\bf Test} &
\multicolumn{2}{c|}{{\bf Vocab Size}} & {\bf Train} &
\multicolumn{2}{c}{{\bf Finetune}}\\
  \cline{5-6} \cline{8-9}
  & {\bf Size}& {\bf Size}& {\bf Size} & Source & Target & {\bf Epoch} & Start & Cycle \\
  \hline
English$\rightarrow$German Translation & 4.5M & 3000 & 3003 & 50K & 50K & 12 & 8 & 1 \\
  \hline
German$\rightarrow$English Translation & 4.5M & 3000 & 2169 & 50K & 50K & 12 & 8 & 1 \\
  \hline
English unsupervised & 12.1M & \multicolumn{2}{c|}{\multirow{2}{*}{Details in
text}} & 50K & 50K & 6 & 4 & 0.5 \\
  \cline{1-2} \cline{5-9}
German unsupervised & 13.8M & \multicolumn{2}{c|}{} & 50K & 50K & 6 & 4 & 0.5 \\
  \hline
Penn Tree Bank Parsing & 40K & 1700 & 2416 & 50K & 104 & 40 & 20 & 4 \\
  \hline
High-Confidence Corpus Parsing & 11.0M & 1700 & 2416 & 50K & 104 & 6 & 4 & 0.5 \\
  \hline
Image Captioning & 596K & 4115 & -  & - & 50K & 10 & 5 & 1 \\ 
\end{tabular}
}
\caption{{\bf Data \& Training Details} -- Information about the different
datasets used in this work. For each task, we display the following
statistics: (a) the number of training examples, (b) the sizes of the
vocabulary, (c) the number of training epochs, and (d) details on when
and how frequent we halve the learning rates ({\it finetuning}).}
\label{t:tasks} 
\end{table}

\subsection{Results}
We explore several multi-task learning scenarios by combining a {\it
large} task (machine translation) with: (a) a {\it small} task -- Penn
Tree Bank (PTB) parsing, (b) a {\it medium-sized} task -- image
caption generation, (c) another {\it large} task -- parsing on the
high-confidence (HC) corpus, and (d) lastly, {\it unsupervised tasks},
such as autoencoders and skip-thought vectors. In terms of evaluation metrics,
we report both validation and test perplexities for all tasks. Additionally, we
also compute test BLEU scores \citep{Papineni02bleu} for the translation task.

\subsubsection{Large Tasks with Small Tasks} 
\label{subsubsec:big_small}

In this setting, we want to understand if a small task such as {\it
PTB parsing} can help improve the performance of a large task such as
translation.  Since the parsing task maps from a sequence of English
words to a sequence of parsing tags \citep{vinyals15grammar}, only the
encoder can be shared with an English$\rightarrow$German translation
task.  As a result, this is a {\it one-to-many}
MTL scenario ($\S$\ref{subsec:otm}).

To our surprise, the results in Table~\ref{t:big_small} suggest that
by adding a very small number of parsing mini-batches (with mixing ratio $0.01$,
i.e., one parsing mini-batch per 100 translation mini-batches), we can improve
the translation quality substantially. More concretely,
our best multi-task model yields a gain of +$1.5$ BLEU points over the
single-task baseline. It is worth pointing out that as shown in
Table~\ref{t:big_small}, our single-task baseline is very strong, even better
than the equivalent non-attention model reported in \citep{luong15attn}. Larger
mixing coefficients, however, overfit the small
PTB corpus; hence, achieve smaller gains in translation quality. 

For parsing, as \citet{vinyals15grammar} have shown that attention is crucial to
achieve good parsing performance when training on the small PTB corpus,
we do not set a high bar for our attention-free systems in this setup (better
performances are reported in Section~\ref{subsub:ll}). Nevertheless, the parsing
results in Table~\ref{t:big_small} indicate that MTL is
also beneficial for parsing, yielding an improvement of up to +$8.9$ F$_1$ points
over the baseline.\footnote{While perplexities correlate well with BLEU scores as shown
in \citep{luong15}, we observe empirically in Section~\ref{subsub:ll} that parsing perplexities are only
reliable if it is less than $1.3$. Hence, we omit parsing perplexities in
Table~\ref{t:big_small} for
clarity. The parsing test perplexities (averaged over two
runs) for the last four rows in Table~\ref{t:big_small} are 1.95, 3.05, 2.14, and 1.66. Valid perplexities
are similar.} 
It would be interesting to study how MTL can be
useful with the presence of the {\it attention} mechanism, which we
leave for future work.

\begin{table}[tbh!]
\centering
\begin{tabular}{l|c|c|c|c}
\multirow{ 2}{*}{\bf{Task}} & \multicolumn{3}{c|}{{\bf Translation}} &
\multicolumn{1}{c}{{\bf
Parsing}}\\
  \cline{2-5}
  & Valid ppl & Test ppl & Test BLEU & Test F$_1$ \\
  \hline
\citep{luong15attn} & - & 8.1 & 14.0 & -  \\
  \hline
\multicolumn{5}{c}{{\it Our single-task systems}} \\
  \hline
Translation & 8.8 (0.3) & 8.3 (0.2) & 14.3 (0.3) & -\\
  \hline
PTB Parsing & - & - & - & 43.3 (1.7) \\
  \hline
\multicolumn{5}{c}{{\it Our multi-task systems}} \\
  \hline
{\it Translation} + PTB Parsing (1x) &  8.5 (0.0) & 8.2 (0.0) & 14.7 (0.1) & 54.5 (0.4) \\
  \hline
{\it Translation} + PTB Parsing (0.1x) &  8.3 (0.1) & 7.9 (0.0) & 15.1 (0.0) &
{\bf 55.2 (0.0)}\\
  \hline
{\it Translation} + PTB Parsing (0.01x) &  {\bf 8.2} (0.2) & {\bf 7.7} (0.2) & {\bf
15.8} (0.4) & 39.8 (2.7) \\
\end{tabular}
\caption{{\bf English$\rightarrow$German WMT'14 translation \& Penn Tree Bank parsing results} --
shown are perplexities (ppl), BLEU scores, and parsing F$_1$ for various systems. For muli-task
models, {\it reference} tasks are in
italic with the mixing ratio in parentheses. Our results are averaged over two
runs
in the format {\it mean (stddev)}. Best results are
highlighted in boldface.}
\label{t:big_small}
\end{table}

\subsubsection{Large Tasks With Medium Tasks} 
We investigate whether the same pattern carries over to a medium task
such as {\it image caption generation}. Since the image caption
generation task maps images to a sequence of
English words \citep{vinyals15caption,xu15}, only the decoder can be
shared with a German$\rightarrow$English translation task. Hence, this
setting falls under the {\it many-to-one} MTL setting ($\S$\ref{subsec:mto}).

The results in Table~\ref{t:big_medium} show the same trend we observed
before, that is, by training on another task for a very small
fraction of time, the model improves its performance on its main task.
Specifically, with 5 parameter updates for image caption generation per 100
updates for translation (so the mixing ratio of $0.05$), we obtain a 
gain of +$0.7$ BLEU scores over a strong single-task baseline. Our baseline is
almost a BLEU point better than the equivalent non-attention model reported in
\cite{luong15attn}.

\begin{table}[tbh!]
\centering
\begin{tabular}{l|c|c|c|c}
\multirow{ 2}{*}{\bf{Task}} & \multicolumn{3}{c|}{{\bf Translation}} &
\multicolumn{1}{c}{{\bf
Captioning}}\\
  \cline{2-5}
  & Valid ppl & Test ppl & Test BLEU & Valid ppl \\ 
  \hline
\citep{luong15attn} & - & 14.3 & 16.9 & - \\ 
  \hline
\multicolumn{5}{c}{{\it Our single-task systems}} \\
  \hline
Translation & 11.0 (0.0) & 12.5 (0.2) & 17.8 (0.1) & - \\ 
  \hline
Captioning & - & - & - & 30.8 (1.3) \\ 
  \hline
\multicolumn{5}{c}{{\it Our multi-task systems}} \\
  \hline
{\it Translation} + Captioning (1x) & 11.9 & 14.0 & 16.7 & 43.3 \\ 
{\it Translation} + Captioning (0.1x) &  10.5 (0.4) & 12.1 (0.4) & 18.0 (0.6) &
{\bf 28.4} (0.3) \\ 
{\it Translation} + Captioning (0.05x) &  {\bf 10.3} (0.1) &  {\bf 11.8} (0.0) &
{\bf 18.5} (0.0) & 30.1 (0.3) \\ 
{\it Translation} + Captioning (0.01x) &  10.6 (0.0) & 12.3 (0.1)& 18.1 (0.4) & 35.2 (1.4)
\\ 
\end{tabular}
\caption{{\bf German$\rightarrow$English WMT'15 translation \& captioning results} -- shown are
perplexities (ppl) and BLEU scores 
for various tasks with similar format as
in Table~\ref{t:big_small}. {\it Reference} tasks are in italic with mixing
ratios in parentheses. The average results of 2 runs are in {\it
mean (stddev)} format.} 
\label{t:big_medium}
\end{table}

\subsubsection{Large Tasks with Large Tasks}
\label{subsub:ll}
Our first set of experiments is almost the same as the one-to-many setting in
Section~\ref{subsubsec:big_small} which combines {\it translation}, as the reference
task, with parsing. The only difference is in terms of parsing data. Instead of using the
small Penn Tree Bank corpus, we consider a large parsing resource, the
high-confidence (HC) corpus, which is provided by \citet{vinyals15grammar}.
As highlighted in Table~\ref{t:big_big_translation}, the
trend is consistent; MTL helps boost translation quality by up
to +$0.9$ BLEU points. 

\begin{table}[tbh!]
\centering
\begin{tabular}{l|c|c|c}
\multirow{ 2}{*}{\bf{Task}} & \multicolumn{3}{c}{{\bf Translation}}\\
  \cline{2-4}
  & Valid ppl & Test ppl & Test BLEU\\
  \hline
\citep{luong15attn} & - & 8.1 & 14.0 \\
  \hline
\multicolumn{4}{c}{{\it Our systems}} \\
  \hline
Translation & 8.8 (0.3) & 8.3 (0.2) & 14.3 (0.3)\\
  \hline
{\it Translation} + HC Parsing (1x) &  8.5 (0.0) & 8.1 (0.1) & 15.0 (0.6) \\
{\it Translation} + HC Parsing (0.1x) &  {\bf 8.2} (0.3) & {\bf 7.7} (0.2) &
{\bf 15.2} (0.6)\\
{\it Translation} + HC Parsing (0.05x) &  8.4 (0.0) & 8.0 (0.1) & 14.8 (0.2) \\
\end{tabular}
\caption{{\bf English$\rightarrow$German WMT'14 translation} -- shown are
perplexities (ppl) and BLEU scores of various translation models. Our
multi-task systems combine translation and parsing on the
high-confidence corpus together. Mixing
ratios are in parentheses and the average results over 2 runs are in {\it
mean (stddev)} format. Best results are bolded.}
\label{t:big_big_translation}
\end{table}

The second set of experiments shifts the attention to {\it parsing} by having it as the reference task. 
We show in Table~\ref{t:big_big_parsing} results that combine parsing with
either (a) the English autoencoder task or (b) the English$\rightarrow$German
translation task. Our models are compared against the best attention-based systems in
\citep{vinyals15grammar}, including the state-of-the-art result of 92.8 F$_1$.

\begin{table}[tbh!]
\centering
\begin{tabular}{l|c|c}
\multirow{ 2}{*}{\bf{Task}}& \multicolumn{2}{c}{{\bf
Parsing}}\\
  \cline{2-3}
  & Valid ppl & Test F$_1$\\
  \hline
  \hline
LSTM+A \citep{vinyals15grammar} &  - & 92.5 \\
LSTM+A+E \citep{vinyals15grammar} & - & {\bf 92.8} \\
  \hline
\multicolumn{3}{c}{{\it Our systems}} \\
  \hline
HC Parsing & 1.12/1.12 & 92.2 (0.1) \\
  \hline
{\it HC Parsing} + Autoencoder (1x) & 1.12/1.12 & 92.1 (0.1) \\
{\it HC Parsing} + Autoencoder (0.1x) & 1.12/1.12 & 92.1 (0.1) \\
{\it HC Parsing} + Autoencoder (0.01x) & 1.12/1.13 & 92.0 (0.1) \\
  \hline
{\it HC Parsing} + Translation (1x) & 1.12/1.13 & 91.5 (0.2) \\
{\it HC Parsing} + Translation (0.1x) & 1.13/1.13 & 92.0 (0.2) \\
{\it HC Parsing} + Translation (0.05x) & {\bf 1.11/1.12} & {\bf 92.4 (0.1)} \\
{\it HC Parsing} + Translation (0.01x) & 1.12/1.12 & 92.2 (0.0) \\
  \hline
Ensemble of 6 multi-task systems & - & {\bf 93.0} \\
\end{tabular}
\caption{{\bf Large-Corpus parsing results} -- shown are
perplexities (ppl) and F$_1$ scores 
for various parsing models. Mixing ratios are in parentheses and the average
results over 2 runs are in {\it mean (stddev)} format. We show the individual perplexities for all runs
due to small differences among them. For \citet{vinyals15grammar}'s parsing results, LSTM+A
represents a single LSTM with attention, whereas LSTM+A+E indicates an ensemble
of 5 systems. Important results are bolded.}
\label{t:big_big_parsing}
\end{table}

Before discussing the multi-task results, we note a few interesting
observations. First, very small parsing perplexities, close to 1.1, can be achieved with large
training data.\footnote{Training solely on the small Penn Tree Bank
corpus can only reduce the perplexity to at most $1.6$, as evidenced by poor
parsing results in Table~\ref{t:big_small}. At the same time, these parsing
perplexities are much smaller than
what can be achieved by a translation task. This is because parsing only has
$104$ tags in the target vocabulary compared to
$50$K words in the translation case. Note that $1.0$ is the theoretical
lower bound.}  
Second, our baseline system can obtain a very competitive F$_1$ score of
92.2, rivaling \citet{vinyals15grammar}'s systems. This is rather surprising
since our models do not use any attention mechanism. A closer look into these
models reveal that there seems to be an architectural difference:
\citet{vinyals15grammar} use 3-layer LSTM with 256 cells and
512-dimensional embeddings; whereas our models use 4-layer LSTM with 1000 cells and
1000-dimensional embeddings. This further supports findings in \citep{rafal16} that
larger networks matter for sequence models.

For the multi-task results, while autoencoder does not seem to help parsing,
translation does. At the mixing ratio of 0.05, we obtain a non-negligible boost of 0.2 F$_1$ 
over the baseline and with 92.4 F$_1$, our multi-task system is on par with the best single system reported in
\citep{vinyals15grammar}. Furthermore, by ensembling 6 different multi-task
models (trained with the translation task at mixing ratios of
0.1, 0.05, and 0.01), we are able to establish a new {\it state-of-the-art} result in
English constituent parsing with {\bf 93.0} F$_1$ score.

\subsubsection{Multi-tasks and Unsupervised Learning}
Our main focus in this section is to determine whether unsupervised
learning can help improve translation. Specifically, we follow the {\it
many-to-many} approach described in Section~\ref{subsec:mtm} to couple
the German$\rightarrow$English translation task with two unsupervised learning
tasks on monolingual corpora, one per language.
The results in Tables~\ref{t:unsupervised_de_en} show a similar trend as before,
a small amount of other tasks, in this case the {\it autoencoder} objective with
mixing coefficient 0.05, improves the translation quality by +$0.5$ BLEU
scores. However, as we train more on the 
autoencoder task, i.e. with larger mixing ratios, the translation performance gets worse. 

\begin{table}[tbh!]
\centering
\resizebox{14cm}{!}{
\begin{tabular}{l|c|c|c|c|c}
\multirow{ 2}{*}{\bf{Task}} & \multicolumn{3}{c|}{{\bf Translation}} &
\multicolumn{1}{c|}{{\bf
German}}& \multicolumn{1}{c}{{\bf English}}\\
  \cline{2-6}
  & Valid ppl & Test ppl & Test BLEU & Test ppl & Test ppl \\
  \hline
\citep{luong15attn} & - & 14.3 & 16.9 & - & -  \\
  \hline
\multicolumn{6}{c}{{\it Our single-task systems}} \\
  \hline
Translation & 11.0 (0.0) & 12.5 (0.2) & 17.8 (0.1) & - & - \\
  \hline
\multicolumn{6}{c}{{\it Our multi-task systems with Autoencoders}}\\
  \hline
{\it Translation} + autoencoders (1.0x) &  12.3 &  13.9 & 16.0 & {\bf 1.01} & 2.10 \\ 
{\it Translation} + autoencoders (0.1x) & 11.4 & 12.7 & 17.7 & 1.13 & {\bf 1.44} \\ 
{\it Translation} + autoencoders (0.05x) & {\bf 10.9} (0.1) & {\bf 12.0} (0.0) &
{\bf 18.3} (0.4) & 1.40 (0.01) & 2.38 (0.39) \\
  \hline
\multicolumn{6}{c}{{\it Our multi-task systems with Skip-thought Vectors}}\\
  \hline
{\it Translation} + skip-thought (1x) & {\bf 10.4} (0.1) & {\bf 10.8} (0.1) & 17.3
(0.2) & {\bf 36.9} (0.1) & {\bf 31.5} (0.4) \\ 
{\it Translation} + skip-thought (0.1x) &  10.7 (0.0) & 11.4 (0.2) & 17.8 (0.4)
& 52.8 (0.3) & 53.7 (0.4) \\ 
{\it Translation} + skip-thought (0.01x) &  11.0 (0.1) & 12.2 (0.0) & {\bf
17.8} (0.3)
& 76.3 (0.8) & 142.4 (2.7) \\ 
\end{tabular}
}
\caption{{\bf German$\rightarrow$English WMT'15 translation \& unsupervised learning results} -- shown are perplexities
for translation and unsupervised learning tasks. We experiment with both {\it
autoencoders} and {\it skip-thought vectors} for the unsupervised objectives. Numbers in {\it
mean (stddev)} format are the average results of 2 runs; others are for 1 run
only.}
\label{t:unsupervised_de_en}
\end{table}

{\it Skip-thought} objectives, on the other hand, behave differently. If we
merely look at the perplexity metric, the results are very encouraging: with
more skip-thought data, we perform better consistently across both the
translation and the unsupervised tasks. However, when computing the BLEU scores,
the translation quality degrades as we increase the mixing coefficients. We anticipate that
this is due to the fact that the skip-thought objective changes the nature of
the translation task when using one half of a sentence to predict the other
half. It is not a problem for the autoencoder objectives, however, since one can
think of autoencoding a sentence as translating into the same language.

We believe these findings pose interesting challenges in the quest towards  better
unsupervised objectives, which should satisfy the following criteria: (a)
a desirable objective should be compatible with the supervised task in focus, e.g.,
autoencoders can be viewed as a special case of translation,
and (b) with more unsupervised data, both intrinsic and extrinsic metrics
should be improved; skip-thought objectives satisfy this criterion in terms of
the intrinsic metric but not the extrinsic one.

\section{Conclusion}
\label{sec:conclude}

In this paper, we showed that multi-task learning (MTL) can improve the
performance of the attention-free sequence to sequence model of
\citep{sutskever14}.  We found it surprising that training on syntactic
parsing and image caption data improved our translation performance, given
that these 
datasets are orders of magnitude smaller than typical translation
datasets. 
Furthermore, we have established a new {\it state-of-the-art} result in
constituent parsing with an ensemble of multi-task models.
We also show that the two unsupervised
learning objectives, autoencoder and skip-thought, behave differently in the MTL context
involving translation. We hope that these interesting
findings will motivate future work in utilizing unsupervised data for sequence
to sequence learning.
A criticism of our work is that our sequence to sequence models do not employ
the attention mechanism \citep{bog15}.  We leave the exploration
of MTL with attention for future work.

\section*{Acknowledgments}
We thank Chris Manning for helpful feedback on the paper and members of the Google Brain team for thoughtful discussions and insights.

\bibliography{iclr16_multi}
\bibliographystyle{iclr16_multi}

\end{document}